\documentclass{article}

\PassOptionsToPackage{numbers, compress}{natbib}

\usepackage[dblblindworkshop, final]{neurips_2025}

\workshoptitle{LAW 2025: Bridging Language, Agent, and World Models}

\newcommand{\cmark}{\textcolor{green!60!black}{\ding{51}}} %
\newcommand{\xmark}{\textcolor{red}{\ding{55}}}            %

\usepackage[backref=page]{hyperref}
\usepackage[utf8]{inputenc} %
\usepackage[T1]{fontenc}    %
\usepackage{url}            %
\usepackage{booktabs}       %
\usepackage{amsfonts}       %
\usepackage{amsmath}        %
\usepackage{nicefrac}       %
\usepackage{microtype}      %
\usepackage{xcolor}         %
\usepackage{graphicx}
\usepackage{subcaption}
\usepackage[titletoc,title]{appendix}
\usepackage[capitalise,nameinlink,noabbrev]{cleveref}
\usepackage{etoolbox}

\AtBeginEnvironment{appendices}{
  \crefalias{section}{appendix}
  \crefalias{subsection}{appendix}
  \crefalias{subsubsection}{appendix}
}

\usepackage{todonotes}
\usepackage{multirow}

\usepackage{pifont}
\usepackage[table]{xcolor}
\usepackage{colortbl}
\usepackage{tabularx}
\usepackage{tikz}

\definecolor{entityblue}{RGB}{2, 255, 255}
\definecolor{roleyellow}{RGB}{255,255,0}
\definecolor{movementgreen}{RGB}{0,255,0}

\DeclareRobustCommand{\highlight}[2]{%
  \tikz[baseline=(X.base)] \node[fill=#1,inner sep=1pt,outer sep=0pt,rounded corners=1pt] (X) {#2};%
}

\usepackage[ruled,vlined]{algorithm2e}
\usepackage{enumitem}
\setlist[itemize]{leftmargin=2em}

\renewcommand*{\backref}[1]{}
\renewcommand*{\backrefalt}[4]{{\footnotesize [%
    \ifcase #1 Not cited.%
	\or Cited on page~#2%
	\else Cited on pages #2%
	\fi%
]}}

\title{Language-conditioned world model improves policy generalization by reading environmental descriptions }

\author{
  Anh (Joe) Nguyen \\
  Oregon State University \\
  \texttt{nguyejoe@oregonstate.edu} \\
  \And
  Stefan Lee \\
  Oregon State University \\
  \texttt{leestef@oregonstate.edu}
}

\begin{document}
\maketitle

\begin{abstract}
    To interact effectively with humans in the real world, it is important for agents to understand language that describes the dynamics of the environment---that is, \textit{how the environment behaves}---rather than just task instructions specifying \textit{what to do}.
For example, a cargo-handling robot might receive a statement like "the floor is slippery so pushing any object on the floor will make it slide faster than usual".
Understanding this dynamics-descriptive language is important for human-agent interaction and agent behavior.
Recent work \cite{dynalang, lwm, reader} address this problem using a model-based approach: language is incorporated into a world model, which is then used to learn a behavior policy.
However, these existing methods either do not demonstrate policy generalization to unseen language or rely on limiting assumptions.
For instance, assuming that the latency induced by inference-time planning is tolerable for the target task or that expert demonstrations are available.
Expanding on this line of research, we focus on improving policy generalization from a language-conditioned world model while dropping these assumptions.
We propose a model-based \textit{reinforcement learning} approach, where a language-conditioned world model is trained through interaction with the environment, and a policy is learned from this model---without planning or expert demonstrations.
Our method proposes \underline{L}anguage-aware \underline{E}ncoder for \underline{D}reamer \underline{W}orld \underline{M}odel (LED-WM) built on top of DreamerV3 \cite{dreamerv3}.
LED-WM features an observation encoder that uses an attention mechanism to explicitly ground language descriptions to entities in the observation.
We show that policies trained with LED-WM generalize more effectively to unseen games described by \textit{novel dynamics and language} compared to other baselines in several settings in two environments: MESSENGER and MESSENGER-WM.
To highlight how the policy can leverage the trained world model before real-world deployment, we demonstrate the policy can be improved through fine-tuning on synthetic test trajectories generated by the world model.

\end{abstract}

\section{Introduction}
\label{sec:introduction}
We envision a future where humans can seamlessly command AI agents through natural language to automate repetitive tasks in the real world.
Traditionally, language has been used to specify task instructions, such as telling a navigation robot to "go to the door" \cite{Bisk2016-hd, Krantz2022-jc, Anderson2017-nb}.
However, language can also offer valuable information about environments.
Such environmental description not only makes human interaction more natural, but also provides important contextual information about how the environment changes over time.
It informs the agent about \textit{how the environment behaves}---its dynamics, the current state of the world, and how various entities interact with each other and with the agent---not just \textit{what to do}.

\begin{figure}
    \centering
    \includegraphics[width=\textwidth]{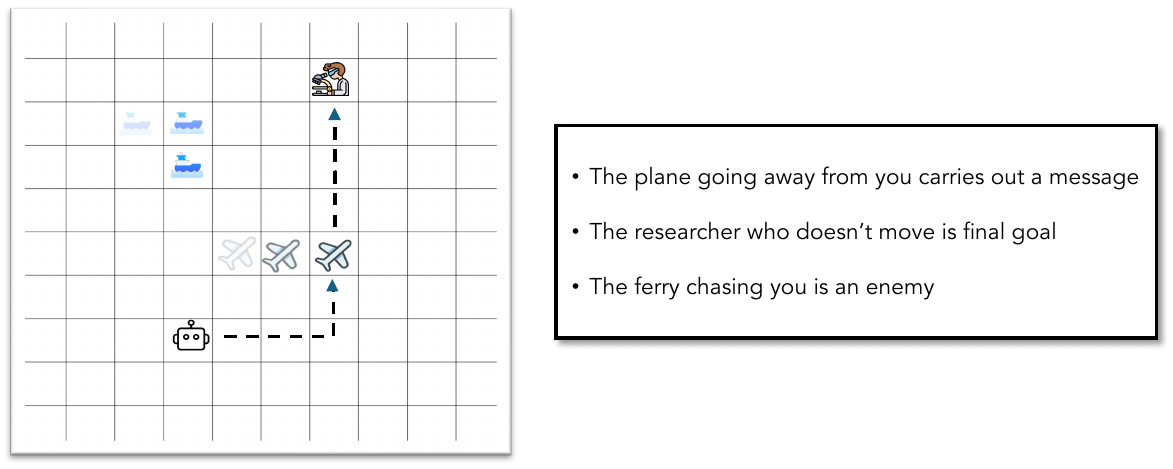}
    \caption{An example of dynamics-descriptive language in a game play.
        The observation includes a 10 $\times$ 10 grid-world with three entities represented by their associated symbols: (\texttt{ferry} - \includegraphics[height=1em]{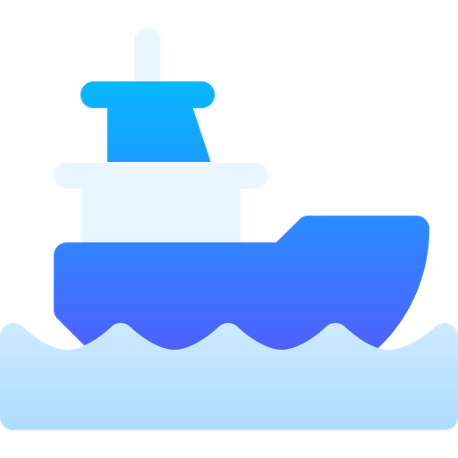}), (\texttt{plane} - \includegraphics[height=1em]{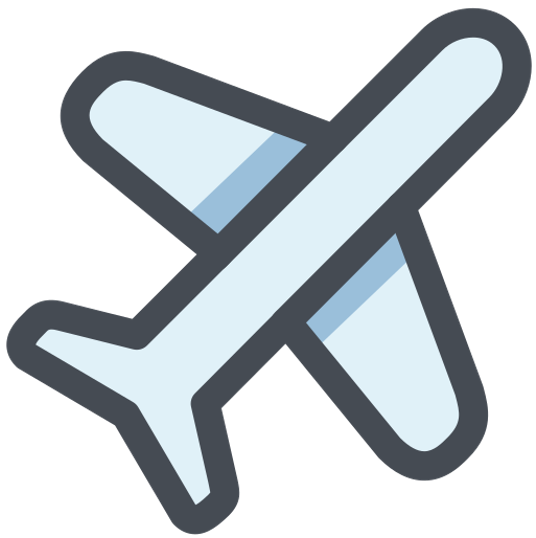}), (\texttt{researcher} - \includegraphics[height=1em]{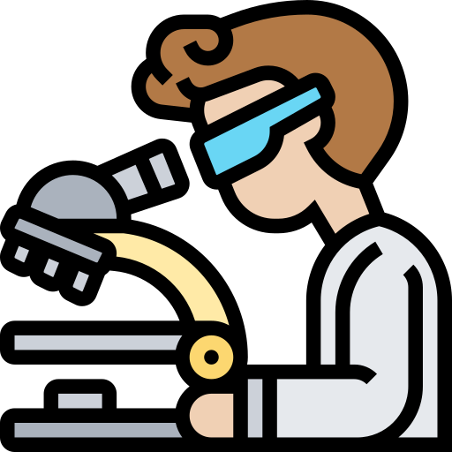}) and one agent (depicted by \includegraphics[height=1em]{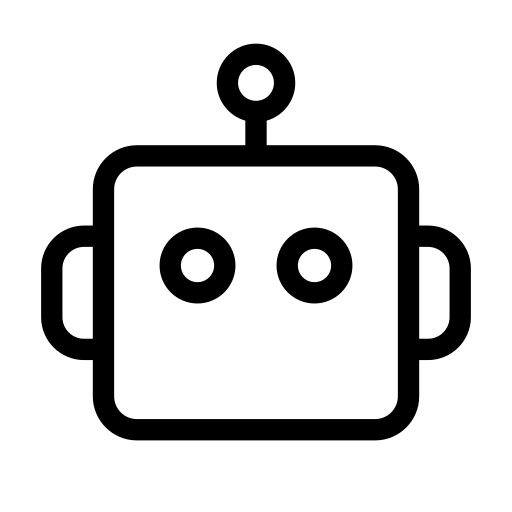}).
        The observation also has a manual on the right, which describes the dynamics of the game.
        The agent can navigate the grid using five actions: \texttt{left}, \texttt{right}, \texttt{up}, \texttt{down}, and \texttt{stay}.
        The agent can only interact with entities when it is in the same grid cell as the entity.
        The agent's task is to identify roles of all entities from the manual, go to the \texttt{messenger}, then go to the \texttt{goal}, while avoiding the \texttt{enemy}.
        Shaded icons indicate one possible scenario of entity movement over time.
        By observing entity movement patterns and grounding language to entities based on their behaviors, the agent can infer the roles assigned to each entity: \texttt{(ferry-enemy)}, \texttt{(plane-messenger)}, and \texttt{(researcher-goal)}.
        The agent can then execute an appropriate plan to complete the task.
        The dashed line in the grid shows such a possible plan.
    }
    \label{fig:messenger_example}
\end{figure}

We illustrate dynamics-descriptive language by using a simple 2D grid-based game in  \cref{fig:messenger_example}, instantiated by MESSENGER S2 \cite{emma}.
This is the setting of our testbed environments and will be detailed in \cref{sec:messenger}
Each game instance consists of several entities, an agent positioned in a grid-world observation, and a language manual.
Each entity has a role among \texttt{messenger}, \texttt{goal}, and \texttt{enemy}.
The agent acts as a courier, tasked with picking up a message from the \texttt{messenger} and delivering it to the \texttt{goal} while avoiding the \texttt{enemy}.
The manual provides descriptions of the entity attributes, helping the agent understand the environment’s dynamics: what the roles of entities are and how the environment changes as the agent interacts with them.
To succeed, the agent must interpret the language manual, identify the entities, and infer their respective roles based on observed behaviors.

Language is valuable because it allows for the description of novel games by recombining known concepts.
For instance, consider \cref{fig:messenger_example} as the training reference game and the following example manual: \texttt{The ship going away from you is the goal you need to go to. The stationary plane is an enemy. The scientist won't move and has an important message.}
The example manual describes an \textit{unseen dynamics} game with known concepts derived from the reference game.
To succeed in this environment---where the dynamics have changed but the rules remains the same---the agent must adopt a different behavior than in the reference game.
This manual also produces \textit{novel surface-level language} through synonyms (e.g. "researcher" vs "scientist") and paraphrases ("won't move" vs "stationary"),

We want to study language grounding and how it affects agent generalizability.
Therefore, we abstract away our observation to a discrete grid-world, thus simplifying perception complexity, similar to existing work \cite{emma, crl, dynalang, lwm, reader}.
Our goal is to develop an agent capable of understanding dynamics-descriptive language by grounding it to discrete entities.
More importantly, we aim for the agent to generalize to \textit{unseen games described by unseen dynamics and/or novel language}, allowing it to adapt agent behavior to new environmental changes.

In the current literature, there are two main approaches to building such an agent: model-free and model-based approach.
Model-free methods \cite{emma, crl} directly map language to a policy.
Language grounding is thus based entirely on policy learning signals, without modeling the environment dynamics.
This might be challenging for agent to learn complex mapping from dynamics-descriptive language to action.
Meanwhile, model-based methods like  EMMA-LWM \cite{lwm}, Reader \cite{reader}, and Dynalang \cite{dynalang} build a world model \cite{Ha2018-nb} simulating trajectories, which are then used to train a policy.
Dynamics-descriptive language is incorporated into the world model, enabling it to use language to predict environmental changes.

However, these existing works have some limitations.
EMMA-LWM requires expert demonstrations---a constraint that may not be always feasible for real-life tasks.
Reader assumes inference-time latency is tolerable for the target tasks.
This is because Reader uses a Monte Carlo Tree Search (MCTS) to look ahead and generate a full plan.
This approach may not be practical for applications that require quick policy responses.
Last, we show that the policy learned from Dynalang fails to generalize over unseen games in \cref{sec:policy_evaluate}.
To address these limitations, we adopt a model-based reinforcement learning (MBRL) approach that builds a language-grounded world model from interaction with the environment, and then use this world model to train a policy.
In contrast to previous methods, our approach does not require expert demonstrations, avoids expensive inference-time planning, and can generalize to unseen games.

We propose \underline{L}anguage-aware \underline{E}ncoder for \underline{D}reamer \underline{W}orld \underline{M}odel (LED-WM), building on a MBRL framework: DreamerV3 \cite{dreamerv3}.
LED-WM introduces a new encoder for DreamerV3 that explicitly grounds entities to their language descriptions, using a simple yet effective attention mechanism.
In this paper, we make the following contributions:
\begin{itemize}
    \item We show that a language-conditioned MBRL without an explicit language grounding to entities, instantiated by Dynalang \cite{dynalang}, fails to generalize over unseen games (see \cref{sec:policy_evaluate}).
    \item By using an attention mechanism in LED-WM to do language grounding, we show that a policy trained from LED-WM can generalize over unseen games better than model-free and model-based baselines in several settings MESSENGER and MESSENGER-WM (see \cref{sec:method}).
    \item We demonstrate that given a trained LED-WM, we can improve a trained policy by fine-tuning it in synthetic test trajectories generated by the world model (see \cref{sec:wm_evaluate}).
\end{itemize}

\section{Background}
\paragraph{Problem formulation.}\label{sec:problem_formulation}
We define our problem as a language-conditioned Markov Decision Process, represented by a tuple with common notations: \((\mathcal{S}, \mathcal{A}, r, T, \gamma, H)\).
$\mathcal{S}$ represents the state space where each state has a 10 $\times$ 10 grid-world observation containing entity symbols and an agent (e.g. \includegraphics[height=1em]{figures/ferry.png}, \includegraphics[height=1em]{figures/plane.png}, \includegraphics[height=1em]{figures/scientist.png}, \includegraphics[height=1em]{figures/bot.png} in \cref{fig:messenger_example}).
Each state also has a language manual $L$, describing \textit{environment dynamics}: transition function \(T(s^\prime|s, a)\) and reward function $r(s,a)$.
$L$ consists of $N$ sentences associated with $N$ entities, where each sentence describes the dynamics of each entity. An example of a state is shown in \cref{fig:messenger_example}.
Action space $\mathcal{A} = \{\texttt{up, down, right, left, stay}\}$ is discrete.
The agent must take a sequence of actions $a_t \in \mathcal{A}$ over a horizon $H$, where time step $t \in [1..H]$, resulting in a state-action trajectory $(s_1, a_1, \ldots, s_H, a_H)$.
Our goal is to find a policy $\pi: \mathcal{S} \times L \rightarrow \mathcal{A}$ that maximizes the expected sum of discounted rewards:
\(
\mathbb{E}_{\pi, L} \left[ \sum_{t=1}^H \gamma^{t-1} r(s_t, a_t) \right].
\)

\paragraph{World model DreamerV3.}
We base our world model on DreamerV3 \cite{dreamerv3}, which uses Recurrent State-Space Model (RSSM) \cite{Hafner2018-db} to build a recurrent world model.
DreamerV3 receives a sequence of observations and predicts latent representations of future observations given actions.
Specifically, at a time step $t$, DreamerV3 receives an observation $x_t$, an action $a_t$, and history information $h_t$.
These inputs are compressed into a latent representation $z_t$ and fed to RSSM with the action $a_t$ to predict the next latent representation $z_{t+1}$.
The world model has the following components:
\begin{align}
    \raisebox{2.5em}{
        $\text{RSSM}~~\begin{cases} \hphantom{A} \\[-6pt] \hphantom{A} \\[-6pt] \hphantom{A} \end{cases}$}
    \begin{alignedat}{3}
         & \text{Sequence model:}     & \quad h_t       & = f_\phi(h_{t-1}, z_{t-1}, a_{t-1})  \\
         & \text{Encoder:}            & \quad z_t       & \sim q_\phi(z_t \mid h_t, x_t)       \\
         & \text{Dynamics predictor:} & \quad \hat{z}_t & \sim p_\phi(\hat{z}_t \mid h_t)      \\
         & \text{Reward predictor:}   & \quad \hat{r}_t & \sim p_\phi(\hat{r}_t \mid h_t, z_t) \\
         & \text{Continue predictor:} & \quad \hat{c}_t & \sim p_\phi(\hat{c}_t \mid h_t, z_t) \\
         & \text{Decoder:}            & \quad \hat{x}_t & \sim p_\phi(\hat{x}_t \mid h_t, z_t) \\
    \end{alignedat}
    \label{eq:dreamerv3}
\end{align}

In this work, we propose to change the encoder of DreamerV3 to better leverage language grounding to learn a more robust world model.

\section{Environment setup}

\label{sec:env}

We adopt MESSENGER \cite{emma} and MESSENGER-WM \cite{lwm} as our test bed environments.
Both environments have the same setup as the example game in \cref{fig:messenger_example}.
To succeed in the game, the agent must understand the language manuals $L$ and use reward and transitional signals to ground the roles, entity names and movement types to the entity symbols in the observation.

\subsection{MESSENGER} \label{sec:messenger}

\paragraph{Overview.}
As shown in \cref{fig:messenger_example}, MESSENGER \cite{emma} is a 10 $\times$ 10 grid-world environment.
We refer the readers to \cref{fig:messenger_example} for game rules and setup, and \cref{sec:env_dynamic_action} for environment dynamics and action.
Each game includes a language manual and an observation containing entities and a single agent.
For more details about language grounding to entities, we refer the readers to \cref{sec:entities_language_manuals}.

\paragraph{Evaluation settings.}
MESSENGER offers four stages (stage S1, S2, S2-dev, S3) with different levels of generalization for test games.
Each stage has its own training set, test set, and development set, all of which are described in detail in \cref{sec:messenger_stages}.

\subsection{MESSENGER-WM}

\paragraph{Overview.}
While providing multiple stages to evaluate policy generalization over out-of-distribution dynamics, MESSENGER does not include a setting for compositional generalization dynamics.
To bridge this gap, MESSENGER-WM \cite{lwm}, derived from MESSENGER S2,
enables evaluation at compositional generalization for world model and policy.
Together, these two environments offer a comprehensive framework for assessing generalization, under varying levels of unseen games.

\paragraph{Evaluation settings.} MESSENGER-WM has three different evaluation settings with different levels of generalization: NewCombo, NewAttr, and NewAll. All settings share the same training set.
More details about the evaluation settings are provided in \cref{sec:messenger_wm_example} and the original paper \cite{lwm}.

\subsection{Evaluating generalization to unseen games}

\begin{table}
    \caption{Summary of generalization capabilities over unseen games in MESSENGER and MESSENGER-WM across stages. Examples with visualizations are provided in \cref{sec:example_gen}.}
    \label{tab:summary}
    \centering
    \small
    \renewcommand{\arraystretch}{1.2}
    \begin{tabularx}{\linewidth}{l *{4}{>{\centering\arraybackslash}X} *{3}{>{\centering\arraybackslash}X}}
        \toprule
                                                        & \multicolumn{4}{c}{\textbf{\scriptsize MESSENGER}} & \multicolumn{3}{c}{\textbf{\scriptsize MESSENGER-WM}}                                                                                                                                                                                                   \\
        \cmidrule(lr){2-5} \cmidrule(lr){6-8}
                                                        & \multirow{2}{*}{\centering\textbf{\scriptsize S1}} & \multirow{2}{*}{\centering\textbf{\scriptsize S2}}    & \multirow{2}{*}{\centering\textbf{\scriptsize S2-dev}} & \multirow{2}{*}{\centering\textbf{\scriptsize S3}} & \textbf{\scriptsize New}   & \textbf{\scriptsize New}  & \textbf{\scriptsize New} \\
                                                        &                                                    &                                                       &                                                        &                                                    & \textbf{\scriptsize Combo} & \textbf{\scriptsize Attr} & \textbf{\scriptsize All} \\
        \midrule
        Novel combinations of known entities            & \cmark                                             & \cmark                                                & \cmark                                                 & \cmark                                             & \cmark                     & \xmark                    & \cmark                   \\
        Novel language (synonyms and paraphrase)        & \cmark                                             & \cmark                                                & \cmark                                                 & \cmark                                             & \cmark                     & \cmark                    & \cmark                   \\
        Novel entity-role assignments                   & \cmark                                             & \cmark                                                & \cmark                                                 & \cmark                                             & \xmark                     & \cmark                    & \cmark                   \\
        Novel entity-movement-role assignments          & \xmark                                             & \cmark                                                & \cmark                                                 & \cmark                                             & \xmark                     & \cmark                    & \cmark                   \\
        Novel game dynamics of known movement behaviors & \xmark                                             & \cmark                                                & \xmark                                                 & \xmark                                             & \xmark                     & \cmark                    & \cmark                   \\
        Novel game dynamics from one training dynamic   & \xmark                                             & \cmark                                                & \xmark                                                 & \xmark                                             & \xmark                     & \xmark                    & \xmark                   \\
        \bottomrule
    \end{tabularx}
\end{table}

\label{sec:generalization_details}
Together, MESSENGER and MESSENGER-WM offer different levels of generalization in test games. We summarize these in \cref{tab:summary} and below:
\begin{itemize}
    \item \textit{Novel language}: the test manual uses synonyms and paraphrases to create novel language through surface structure.
    \item \textit{Novel combinations of known entities}: the test game involve entities that appear in training set but never appear together in one training game.
    \item \textit{Novel entity-role assignments}: at least one entity in the test game has a different role from its roles in training games.
    \item \textit{Novel entity-role-movement assignment}: at least one entity in the test game has a novel combination of entity-role-movement assignment.
    \item \textit{Novel combinations of known movement behaviors (novel game dynamics)}: the test game has a novel movement combinations of entities, e.g. \texttt{(chaser-chaser-chaser)} for three entities.
    \item \textit{Novel game dynamics from one training dynamic}:
          Game dynamic is defined by the combination of entity movements.
          In the training set, there is only one such combination \texttt{(chasing-fleeing-stationary)} across all training games.
          The test game meanwhile has a novel dynamic, e.g. \texttt{(chaser-chaser-chaser)}.
          This is also the difference between MESSENGER-WM and MESSENGER, which can be found more detailed in \cref{sec:diff_wm_messenger}
\end{itemize}%
See \cref{sec:example_gen} for visualizations of these settings.

\section[Method: LED-WM]{Method: \underline{L}anguage-aware \underline{E}ncoder for \underline{D}reamer \underline{W}orld \underline{M}odel (LED-WM)}
\label{sec:method}
\begin{figure}
    \centering
    \includegraphics[width=\textwidth]{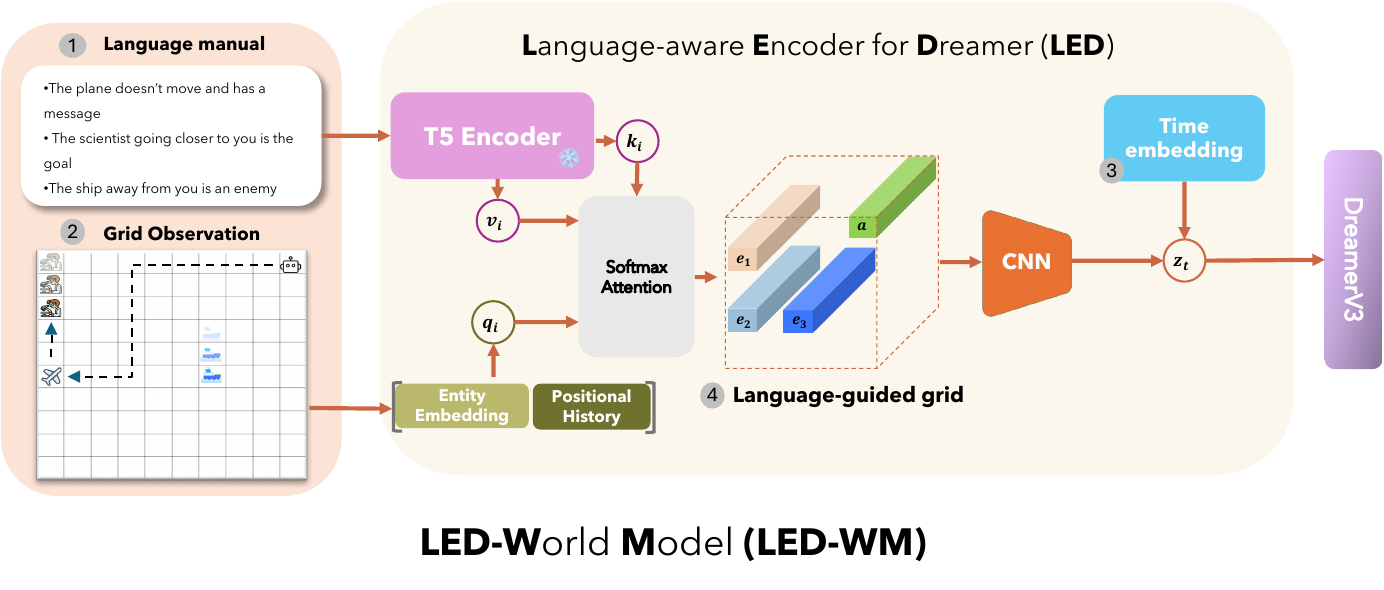}
    \caption{
        Overview of our proposed world model LED-WM.
        The world model input consists of: \includegraphics[height=1em]{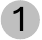} a language manual \( L \), \includegraphics[height=1em]{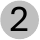} a grid-world observation representing entity and agent symbols, and \includegraphics[height=1em]{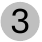} the current time step \( t \).
        Entity, agent symbols, and time step are encoded using learned embeddings, while \( L \) is encoded via a frozen T5 encoder.
        To represent each entity, we employ a multi-layer perceptron (MLP) that processes the entity embedding and its temporal information, capturing its movement pattern relative to the agent, to produce a query vector.
        We apply an attention network  between the query vectors and the sentence embeddings to align each entity with its corresponding sentence.
        The resulting vectors are then put into their respective entity positions.
        This produces \includegraphics[height=1em]{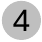} a language-grounded grid \( G_l \), which is then processed by a CNN.
        The extracted feature vector is flattened and concatenated with the time embedding to form final observation representation \( x_t \).}
    \label{fig:led}
\end{figure}

To generalize policy across unseen games, we aim to develop a world model capable of doing language grounding to entities in a game.
Inspired by EMMA \cite{emma}, we propose \underline{L}anguage-aware \underline{E}ncoder for \underline{D}reamer (LED), which uses cross-modal attention to align game entities with sentences.
The resulting vectors are then placed back into their original entity locations, producing a language-aware grid observation.
This grid is passed through a CNN encoder to extract observation features, which are used by the other components of DreamerV3.
We call this overall model LED-\underline{W}orld \underline{M}odel (LED-WM). We provide an overview of the encoder LED and the world model LED-WM in \cref{fig:led} and describe each component in the following sections. 

\subsection{Observational inputs}

The input to the world model consists of a natural language manual \( L \), a grid observation \( o_t \) of size $10 \times 10$, containing symbolic entities, and the current time step \( t \).
The manual \includegraphics[height=1em]{figures/1.pdf} comprises \( N \) sentences, each describing the dynamic of one of the \( N \) entities in the observation.
Following \citet{dynalang}, \( N \) sentences in \( L \) are encoded using a T5 encoder \cite{Raffel2023-ru}, resulting in \( N \) frozen sentence embeddings, denoted by \( s_1, s_2, ..., s_N\).
In the grid \includegraphics[height=1em]{figures/2.pdf}, the \( N \) entities and the agent, represented by entity symbols, are encoded using a learned entity embedding vectors initialized with random weights.
\footnote{This ensures the agent does not have prior knowledge about entity identities, requiring it to infer entities based on the language.}
This results in $N$ symbol embeddings $sb_1, sb_2, ..., sb_N \in \mathbb{R}^{d_{sb}}$ and a single agent embedding $a \in \mathbb{R}^{d_{sb}}$.
The current time step \( t \) \includegraphics[height=1em]{figures/3.pdf} is encoded as $time_t$ using a learned time embedding, also initialized with random weights.

To build position history of each entity \( i \), we capture temporal dynamics by constructing an array \( D_i \) temporally, with length corresponding to the maximum possible steps in the environment and initial values of \(-1\).
At time step \( t \), let the 2D coordinate of the entity \( i \) be \( p^t_i \) and that of the agent be \( p^t_a \).
To determine the relative direction of the entity’s movement with respect to the agent, we compute the dot product:
\begin{equation}
    D_i^t = \frac{p^t_i - p^t_a}{\|p^t_i - p^t_a\|} \cdot \frac{p^t_i - p^{t-1}_i}{||p^t_i - p^{t-1}_i||}, \forall i\in[1..N], \forall t,
\end{equation}
where the first term is a normalized vector from the agent to the entity \( i \), and the second term is a normalized velocity vector of the entity \( i \).
This dot product quantifies the alignment between the entity’s direction of motion and its position relative to the agent at each time step \( t \).

\subsection{LED: Building a language-aware encoder} \label{sec:led}
We construct a language-grounded grid representation that aligns the language manual $L$, which consists of $N$ sentence embeddings, with the observation $o_t$, which includes $N$ entity embeddings and one agent embedding.
To align the sentence embeddings with the entity embeddings, we use an attention network.
The values are obtained through a linear transformation of the sentence embeddings $s_i$.
Meanwhile, the queries are obtained through a multi-layer perception (MLP) applied to the entity embeddings $sb_i$ and temporal array $D_e$.
Likewise, the keys are obtained through an MLP applied to the sentence embeddings $s_i$ :
\begin{align}
    q_i & = \text{MLP}([sb_i, D_e]), & k_i & = \text{MLP}(s_i), & v_i & = W_v s_i,                       \\
    q_i & \in \mathbb{R}^d,          & k_i & \in \mathbb{R}^d,  & v_i & \in \mathbb{R}^{d_{\text{val}}},
\end{align}
where $d$ and $d_{val}$ denote the dimensions of the query/key and value vectors, respectively.
We then apply scaled dot-product attention \cite{Vaswani2023-iq}. Given $K \in R^{N \times d} $ as the key matrix where the row $i$ of $K$ is $k_i^T$, attention scores $\gamma_i \in R^N$ and resulting vector $e_i \in R^{d_a}$ for each entity $i$ are calculated as:
\begin{align}
    \gamma_i = softmax \left ( \frac{q_i \cdot K  }{\sqrt{d}} \right ), \quad e_i = \sum_{j=1}^N \gamma_{ij} v_j,
\end{align}
This attention aligns entity symbols in the observation with sentences in the manual based on attribute language descriptions such as movement (e.g., \texttt{chaser, moving away, stationary}) and entity name (e.g., \texttt{dog, wizard}).
The resulting $e_i$ from the attention is able to represent an associated role for entity $i$ such as \texttt{enemy, messenger, goal}, which is vital information for world model and policy learning.

To retain the spatial information of entities, we place the resulting vectors \( e_i \) back into the original positions of their corresponding entities in the grid observation.
This produces \includegraphics[height=1em]{figures/4.pdf} a language-aware grid observation \( G_l \) of size \( h \times w \times d_{val} \).
We then use a CNN encoder to extract a feature map, which is subsequently flattened and concatenated with the time embedding \( \text{time}_t \).
The combined representation is processed through an MLP to obtain the final feature representation \( x_t \) for the observation \( o_t \) at time step \( t \):
\begin{align}
    x_t & = \text{MLP}(\text{Flatten}([\text{CNN}(G_l)), \text{time}_t])
\end{align}

Denoting $\phi$ as the parameters of LED-WM, we can find stochastic variable $z_t$ as the function of $x_t$:
\begin{align}
    z_t \sim q_\phi(z_t |h_t, x_t),
\end{align}
which now replaces the encoder in DreamerV3, as shown in \cref{eq:dreamerv3}.

\subsection{LED-WM: Combining LED with Dreamerv3}

We replace DreamerV3's encoder with LED, resulting in our world model LED-WM.
We adopt world model and policy learning from DreamerV3.
However, we make the following changes to the original architecture to improve policy generalization and sample efficiency: we omit the reconstruction decoder (Decoder in \cref{eq:dreamerv3}) and adopt multi-step prediction for reward and continue prediction \cite{Hansen2023-cd, Peri2024-kb}.
For more details, we refer the readers to \cref{sec:led_wm_details} for world model loss and \cref{sec:training_procedure} for training procedure.

\section{Experiments}
\label{sec:experiment}
We want to answer the following questions: 1) Can a policy trained on our world model LED-WM generalize to unseen games? (see \cref{sec:policy_evaluate}),
and 2) Can the world model LED-WM generalize to unseen games? (see \cref{sec:wm_evaluate})
To answer these, we use two environments: MESSENGER and MESSENGER-WM, which are detailed in \cref{sec:env}.
We detail the training settings in \cref{sec:training_details}.

\subsection{Policy generalization trained from LED-WM} \label{sec:policy_evaluate}

\subsubsection{Policy baselines} \label{sec:policy_baselines}
As baselines, we adopt the following model-free (EMMA,CRL) and model-based (Dynalang, EMMA-LWM) methods:
\begin{itemize}
    \item \textit{EMMA} \cite{emma} uses attention between entities and sentences to generate language-conditioned observation to the policy.
          The policy is trained via curriculum learning where the agent is initialized with parameters learned from previous easier game settings.
          We report EMMA with curriculum learning from the original paper and EMMA without curriculum learning from \cite{lwm}.

    \item \textit{CRL} \cite{crl} develops a specialized constraint for MESSENGER to overcome spurious correlations between entity identities and their roles in the training data.
          It has the state-of-the-art win rate performance in test environments of MESSENGER.
    \item    \textit{Dynalang} \cite{dynalang} use soft actor-critic for policy learning.
          Because the paper does not report policy generalization performance in MESSENGER, we first reproduce Dynalang using published code and train to convergence according to published hyperparameters and training steps.
          We then report its policy performance on test environment of MESSENGER in \cref{tab:policy_messenger}.

    \item \textit{EMMA-LWM} \cite{lwm} built a language-conditioned world model.
          A policy is trained with simulated trajectories from this world model through online imitation learning and filtered behavior cloning.
          Both methods require expert demonstrations.
          \footnote{Online imitation learning is where the expert supervises the optimal action to take in simulated states (from the world model). Meanwhile, in filtered behavior cloning, the expert uses only states from its own expert plan. The agent then only chooses plans that achieve the highest returns according to the world models to imitate.}

\end{itemize}

\subsubsection{Evaluation metrics} \label{sec:policy_evaluation_metrics}
\begin{itemize}
    \item \textit{Win Rates for MESSENGER}: To make our comparison consistent with reported results from EMMA \cite{emma} and CRL \cite{crl}, we adopt win rate as the metric in MESSENGER.
          Win rate is calculated as the average number of games won by the agent over 1000 episodes.

    \item \textit{Average Sum of Scores for MESSENGER-WM}: Likewise to be consistent with EMMA-LWM \cite{lwm} studying MESSENGER-WM, we adopt average sum of scores as the metric.
          For each game configuration, we run the policy for 60 trials \footnote{We find that 60 trials are enough to find a stable average sum of scores to evaluate a policy given a particular game configuration. } and compute the average sum of scores.
          This process is repeated for 1000 games, and we report the average sum across all games.

\end{itemize}

\begin{table}
    \centering
    \small
    \renewcommand{\arraystretch}{1.3}
    \caption{Policy generalization in MESSENGER in terms of win rate.
        Note that other methods (Dynalang, CRL and LED-WM) do not use curriculum training.
        Results of Dynalang and LED-WM ($^*$) are rounded to second decimal place, while results for CRL and EMMA are taken from their original papers.
        Results are recorded across five training seeds.}

    \label{tab:policy_messenger}
    \begin{tabular}{lcccc}
        \hline
        \multirow{2}{*}{\parbox[c]{2cm}{\centering\textbf{Method}}} & \multicolumn{4}{c}{\textbf{MESSENGER}}                                                                 \\ \cline{2-5}
                                                                    & \textbf{S1}                            & \textbf{S2}     & \textbf{S2-dev}     & \textbf{S3}           \\
        \midrule
        Dynalang$^*$                                                & 0.03 ± 0.02                            & 0.04 ± 0.05     & --                  & 0.03 ± 0.05           \\
        CRL                                                         & 88 ± 2.5                               & \textbf{76 ± 5} & --                  & 32 ± 1.9              \\
        EMMA (w/o curriculum)                                       & 85 ± 1.4                               & 45 ± 12         & --                  & 10 ± 0.8              \\
        EMMA(w/ curriculum)                                         & 88 ± 2.3                               & 95 ± 0.4        & --                  & 22 ± 3.8              \\
        \midrule
        \textbf{LED-WM (Ours)}$^*$                                  & \textbf{100 ± 0}                       & 51.6 ± 2.7      & \textbf{96.6 ± 1.0} & \textbf{34.97 ± 1.73} \\
    \end{tabular}
\end{table}

\subsubsection{Results} \label{sec:policy_results}

We report the win rate performance of our method and other baselines for MESSENGER in \cref{tab:policy_messenger} and the average sum of scores for MESSENGER-WM in \cref{tab:policy_messenger_wm}.

In MESSENGER-WM, LED-WM outperforms EMMA-LWM in all settings without using any expert demonstrations.
In MESSENGER, Dynalang fails to generalize to unseen games. We hypothesize that this is because Dynalang lacks an explicit mechanism to ground language to each entity. Meanwhile, LED-WM is better than other baselines in S1 and comparable to CRL in S3.

However, LED-WM underperforms CRL in S2, where the agent is trained on only one movement combination \texttt{chasing-fleeing-stationary} but is evaluated over different unseen movement combinations (unseen dynamics - see \cref{tab:summary}).
In contrast, LED-WM performs well on S2-dev, where its setting is similar to S2, but its test dynamics are the same as the training games.
We hypothesize that this occurs because CRL incorporates an explicit mechanism to mitigate the data bias in S2 that there is only one movement combination in the training data and spurious correlations between entity identities and their roles.
For instance, the assumption "a dog is always a goal".
Therefore, this mechanism might enhance generalization in test scenarios where the dog is either a friend or an enemy.
Incorporating such a mechanism in LED-WM might be a promising direction for future work.

\begin{table}
    \centering
    \small
    \renewcommand{\arraystretch}{1.3}
    \caption{Policy generalization in MESSENGER-WM in terms of average sum of scores. EMMA-LWM results are taken from its original paper \cite{lwm}. Results are recorded across five training seeds.}
    \label{tab:policy_messenger_wm}
    \begin{tabular}{lcccc}
        \hline
        \multirow{2}{*}{\parbox[c]{3cm}{\centering\textbf{Method}}} & \multicolumn{3}{c}{\textbf{MESSENGER-WM}}                                                        \\ \cline{2-4}
                                                                    & \textbf{NewCombo}                         & \textbf{NewAttr}         & \textbf{NewAll}           \\
        \hline
        EMMA-LWM                                                    &                                           &                          &                           \\
        \hspace{0.5cm} Online IL                                    & 1.01 $\pm$ 0.12                           & 0.96 $\pm$ 0.17          & 0.62 $\pm$ 0.21           \\
        \hspace{0.5cm} Filtered BC (near-optimal)                   & 1.18 $\pm$ 0.10                           & 0.75 $\pm$ 0.20          & 0.44 $\pm$ 0.18           \\
        \hspace{0.5cm} Filtered BC (suboptimal)                     & 0.98 $\pm$ 0.13                           & 0.29 $\pm$ 0.25          & 0.13 $\pm$ 0.19           \\ \hline
        \textbf{LED-WM (Ours)}                                      & \textbf{1.31 $\pm$ 0.05}                  & \textbf{1.15 $\pm$ 0.08} & \textbf{1.16 $\pm$ 0.02 } \\
    \end{tabular}
\end{table}

\subsection{World model generalization} \label{sec:wm_evaluate}
To evaluate the generalization of a world model, one pragmatic metric is to measure how its generated rollouts on unseen dynamics benefit policy learning.
If the world model can generalize to unseen dynamics in test games, which effectively simulates these dynamics, a policy finetuned on these rollouts should improve in new games.

\paragraph{Finetuning procedure.}
Given a trained LED-WM, a trained policy from LED-WM, and a test game, LED-WM takes the initial observation and the manual as input to generate 60 synthetic trajectories.
These trajectories are then used to determine whether the policy should be finetuned on this game.
We estimate the value of the trained policy from the world model and finetune the policy if the estimated value is smaller than a pre-defined threshold.
For each gradient update on the policy finetune, we generate 60 synthetic trajectories.
We repeat this process in 2000 optimization steps.
We illustrate the finetuning procedure in \cref{sec:finetune_algo} with a Python-like format.

\begin{table}
    \centering
    \small
    \renewcommand{\arraystretch}{1.3}
    \caption{World model generalization over S2-dev and S3 (MESSENGER) through finetune procedure. Finetune results are recorded in average sum of scores across five seeds.}
    \label{tab:finetune_messenger}
    \begin{tabular}{lcccc}
        \hline
        \textbf{Method} & \textbf{S1}   & \textbf{S2} & \textbf{S2-dev}   & \textbf{S3}       \\
        \hline
        LED-WM (Ours)   & 1.500 $\pm$ 0 & -           & 1.4478 $\pm$ 0.01 & --0.11 $\pm$ 0.05 \\
        \hline
        After finetune  & -             & -           & 1.4513 $\pm$ 0.01 & -0.01 $\pm$  0.12 \\
    \end{tabular}
\end{table}

\paragraph{Evaluation metrics and results.}
We adopt the average sum of scores due to its robustness to the stochasticity of the environment.
We show policy finetune results in \cref{tab:finetune_messenger} for MESSENGER.
In MESSENGER, we show that the finetuning procedure improves the trained policy in S2-dev,
\footnote{In S2-dev, we use Wilcoxon signed-rank \cite{Wilcoxon1945-sa} and hierarchical bootstrap sampling \cite{Davison2013-hr} corresponding to two levels of hierarchies in our experiments (episodes and run trials): at a 95\% confidence level, hierarchical sampling indicates an improvement between 0.014 and 0.019.} and in S3, demonstrating that the world model is generalizable to test trajectories. However, the absolute policy improvement is still limited in our experiments.

\section{Related work}
Due to space limit, we provide a detailed related work in \cref{sec:related_work_detail}. In this section, we briefly review related work on \textbf{language-conditioned dynamics using model-based approach}.
Recent efforts focus on integrating dynamics-descriptive language into world models, resulting in language-conditioned world models.
Dynalang \cite{dynalang} shows that such world model improves policy's sample efficiency compared to model-free approaches.
However, it does not demonstrate policy generalization in unseen games.
Reader \cite{reader} shows that a MCTS planner can generalize to unseen games using a language-conditioned world model.
Despite this, its environment (RTFM \cite{Zhong2021-sx}) does not require language grounding to entities.
\citet{lwm} introduce MESSENGER-WM, a compositional benchmark based on MESSENGER, and EMMA-LWM---a policy can generalize over unseen games from a language-based world model and expert demonstrations.
Though sharing this same goal of policy generalization with our work, these studies rely on limiting assumptions.
Planning with an MCTS tree in Reader, involves incurring computational cost to generate plans in inference time.
This approach may not be practical for applications that require quick policy responses.
On the other hand, EMMA-LWM requires expert demonstrations to use imitation learning and behavior cloning.
This assumption may not always be feasible for every application.
In contrast, our work lifts these assumptions and demonstrates policy generalization over unseen games in two environments that require language grounding: MESSENGER and MESSENGER-WM.

\section{Conclusion}
We develop an agent that can understand dynamics-descriptive language in interactive tasks.
We adopt a model-based reinforcement learning (MBRL) approach, where a language-conditioned world model is trained through interactions with the environment, and a policy is learned from this world model.
Unlike existing works, we do not require expert demonstrations or expensive planning during inference.
Our method proposes Language-aware Encoder for Dreamer World Model (LED-WM).
LED-WM adopts an attention mechanism to explicitly align language description to entities in the observation.
We show that policies trained with LED-WM can generalize better to unseen games than existing baselines.
We can also further improve the trained policy through fine-tuning on synthetic test trajectories generated by the world model.

\section{Acknowledgment}
We thank everyone from VIRL lab (Oregon State University), especially Skand and Akhil for their valuable feedback and discussions.
The first author was personally supported by Amanda Putiza, Nguyen Thi Ngoc Anh, Ngo Thi Bich Lan, Tran Thanh Nhu, Bui Thuy Tien, and Nguyen Hoang Kieu Anh.
This work is supported by NSF CAREER Award 2339676. We also thank the anonymous reviewers for their valuable feedback and suggestions.

\bibliographystyle{plainnat}
\bibliography{paperpile}

\begin{appendices}
    \newpage

\section{Training details}
\label{sec:training_details}

\begin{table}[h]
    \centering
    \setlength{\abovecaptionskip}{10pt}
    \begin{tabular}{ll}
        \toprule
        Hyperparameter                 & Value \\
        \midrule
        Batch size                     & 30    \\
        Batch length                   & 300   \\
        Optimizer                      & Adam  \\
        World model learning rate      & 3e-4  \\
        Max. world model gradient norm & 30    \\
        Actor learning rate            & 2e-4  \\
        Max. actor gradient norm       & 100   \\
        Critic learning rate           & 1e-4  \\
        Max. critic gradient norm      & 100   \\

        \bottomrule
    \end{tabular}
    \caption{Training hyperparameters.}
    \label{tab:training_hparams}
\end{table}

\begin{table}[h]
    \centering
    \setlength{\abovecaptionskip}{10pt}
    \begin{tabular}{lcc}
        \toprule
        Hyperparameter             & Symbol        & Value \\
        \midrule
        Dynamics loss scale        & $\beta_{dyn}$ & 1     \\
        Representation loss scale  & $\beta_{rep}$ & 0.1   \\
        Latent unimix              & ---           & 1\%   \\
        Free nats                  & ---           & 1     \\
        Sentence embedding dim     & $d_s$         & 32    \\
        Symbol/Agent embedding dim & $d_{sb}$      & 32    \\
        MLP layers                 & ---           & 3     \\
        MLP hidden units           & ---           & 512   \\
        Query/key dim              & d             & 128   \\
        Value dim                  & $d_{val}$     & 128   \\
        RSSM deterministic dim     & ---           & 512   \\
        \bottomrule
    \end{tabular}
    \caption{World model hyperparameters. Other hyperparameters are the same as in DreamerV3 \cite{dreamerv3}. }

    \label{tab:wm_hparams}
\end{table}

\begin{table}[h]
    \centering
    \setlength{\abovecaptionskip}{10pt}
    \begin{tabular}{lcccccc}
        \toprule
        Hyperparameter             & Symbol  & S1 & S2  & S2-dev & S3  & MESSENGER-WM \\
        \midrule
        Number of entities         & N       & 3  & 3   & 3      & 5   & 3            \\
        Episode horizon            & H       & 4  & 32  & 32     & 32  & 32           \\
        Finetune threshold         & $thres$ & -  & 1.2 & 1.2    & 1.4 & -            \\
        Training environment steps & -       & 1M & 10M & 10M    & 20M & 10M          \\
        Training GPU hours         & -       & 6  & 24  & 24     & 72  & 24           \\
        \bottomrule
    \end{tabular}
    \caption{Environment hyperparameters. Training GPU hours are estimated based on 1 NVIDIA H100 GPU.}

    \label{tab:env_hparams}
\end{table}

\section{Detailed related work}
\label{sec:related_work_detail}

\paragraph{How language is used in RL tasks?}

Language is often employed as step-by-step instructions or goal specification in domains such as 1) visual language navigation (VLN) \cite{Krantz2022-gg}
\cite{Krantz2023-ro} \cite{Anderson2017-nb} 2) grid-world games like BabyAI \cite{Chevalier-Boisvert2019-bf} and SILG benchmark \cite{Zhong2022-tw}, and 3) manipulation tasks \cite{Ren2023-jp} \cite{Zheng2023-uh} \cite{Parakh2023-ey}.
Another research direction in language for RL explores how language can accelerate policy learning by providing richer feedback rather than just numerical rewards: language for plan correction \cite{Sharma2022-nf} \cite{Liu2023-lr} \cite{McCallum2023-pb} \cite{Cheng2023-bz} \cite{Nguyen2022-gv} \cite{Mehta2024-jm}, providing more descriptions of the current state or current goal \cite{Nguyen2022-gv} \cite{dynalang}, generating dense rewards \cite{Goyal2019-yd} \cite{Yu2023-tk} \cite{Xie2023-mw}, clarifying information \cite{Mehta2024-jm}, and speeding up exploration \cite{Tam2023-kf}.
This study investigates an alternative use of language in RL problems: describing the dynamics of environments.

\paragraph{Language-conditioned dynamics environments.}

In language-conditioned environments, while language can be used as step-by-step instructions \cite{Krantz2022-jc} \cite{Bisk2016-hd} \cite{Anderson2017-nb}, language can also be used to describe environment dynamnics---that is, \textit{how environments change over time}.
Formally, language describes the transition function $T(s^\prime|s, a)$ and reward function $r(s,a)$ of a MDP system defined in \cref{sec:problem_formulation}.
Several environments have been proposed to provide language-conditioned dynamics \cite{Cao2023-ed} \cite{Zhong2021-sx} but their settings do not require understanding entity interaction or grounding language to entities.
To fill this gap, as discussed in \cref{sec:env} about the environment setup, \citet{emma} present MESSENGER, a more challenging game requiring language grounding to entities based on environment dynamics.
\citet{lwm} later propose MESSENGER-WM built out of MESSENGER to test compositional generalization world model and policy.
In this study, we focus on MESSENGER and MESSENGER-WM due to their requirements for language grounding to entities and a history of previous works on these datasets.

\paragraph{How to solve language-conditioned dynamics environments?}
Generally speaking, there are two ways to understand language-conditioned dynamics: model-free and model-based learning.
First, in model-free approach, language is used to build language-conditioned observation, which is then fed directly to policy in a model-free manner.
Second, in model-based learning, language is used to build language-conditioned world model, which is then used to plan or learn policy.
Our work focuses on the second category: we use dynamics-related language explicitly for dynamics learning, in model-based RL fashion, thus improving policy performance over model-free approaches.

\paragraph{Language-conditioned dynamics in RL: model-free approach.}

In model-free approach, language is used to build language-conditioned observation, which is then fed directly to policy.
\citet{emma} utilize an attention mechanism to ground language to individual entities, forming a language-aware representation for the policy.
\citet{Zhong2022-km} develop a model of environmental dynamics learned from language-conditioned and state-only (without action) demonstrations.
This dynamics model is then used to initialize and distill to the representation of a policy learner, which helps sample-efficiency and generalization across language RL tasks.
\citet{Wang2021-fc} proposes compact and invariant concept-like representations through extracting similarities across observations, which is then proved to be useful for policy learning.
\citet{Wang2021-fc} proposes two-agent system where the manager agent reads the instructions and manuals to devise the plan with sub-goals and the worker agent fulfills the sub-goals in the plan one-by-one.
The model, however, assumes access to the sub-goal text instructions to train the manager.
Those works use language directly for policy learning while our work uses language for world model learning.

\paragraph{Language-condition instruction-based world model (LWM).}

World models in model-based reinforcement learning (RL) involve learning the dynamics of the environment.
In visual-understanding interaction domains like robotics and video games, world models have been widely studied and are empirically proven to be sample-efficient for policy learning \cite{Kaiser2020-lo} \cite{dreamerv3} \cite{Peri2024-kb}.
However, in language-understanding interaction tasks, most language-conditioned world models have been developed to process task instructions or action descriptions, rather than to capture environmental dynamics.
\citet{Poudel2023-jy} integrates human language into the world model, however language primarily describes observations rather than environment's dynamics.
For example, a description like "there is an apple on the left, 2 meters from here" describes the future observation of environment without addressing the transition function.
Recent works \cite{Zhou2024-xk} \cite{Du2023-st} \cite{Zhang2024-uc} develop LWM with compositional generalizability.
While these works use more visually realistic input and require more reasoning to solve their tasks, the language they study is task instruction that involves straightforward mapping from language to objects such as colors and object names, e.g. "Move A Red Block to A Brown Box."

To bridge this gap, we focus on language-conditioned world models that incorporate dynamics-descriptive language—language that explains how entities interact and how the environment changes—rather than just providing direct task instructions.
In contrast, the language used in our testbed environments MESSENGER and MESSENGER-WM describes  environmental dynamics, which is the focus of our study.

\paragraph{Large language model (LLM) for world model.}

Recent work \cite{Dainese2024-zd} \cite{Tang2024-wy} \cite{Piriyakulkij2025-ve} \cite{Zhou2025-zs} use LLMs to build a world model for policy learning.
However, their environments does not require language understanding or language grounding. In other words, they do not build \textit{a language-conditioned world model} like our work.
Further, they do not study the generalization behavior of LLM-based world model in a out-of-distribution (OOD) set up.
In contrast, our work focuses on building a world model that requires language grounding to entities.
We also study the generalization behavior of a language-conditioned world model and a policy learned from this world model.
We achieve this by running experiments in a controlled OOD set up from MESSENGER and MESSENGER-WM.
\section{Environment details}
\subsection{MESSENGER}  \label{sec:messenger_stages}

\subsubsection{Environment Dynamics and Action} \label{sec:env_dynamic_action}
\paragraph{Reward and Game Ending.} The agent loses the game and incurs -1 reward
\footnote{In the $S3$ setting of the game, there is an inconsistency in the environment implementation with the description provided in \cite{emma}: when the agent collides with two enemy entities, the environment returns a reward of -2 instead of -1.
    We observe that this rarely happens and thus has no significant impact on expected policy's behavior. } if either of two events occurs - the agent is in the same cell as the enemy or reaches the goal without first getting the message.
Reaching the messenger gives the agent a reward of 0.5, and then reaching the goal provides a reward of 1.

\paragraph{Observation change.} The agent can be with or without a message, represented by two different symbols in the observation.
The observation changes when the agent interacts with entities according to their roles, specifically:
\begin{itemize}
    \item When the agent without a message picks up \texttt{messenger}, the messenger disappears from the observation. The agent now has a message and is represented by a different symbol from when it was without a message.
    \item When the agent loses the game, either by reaching the goal without first getting the message or by touching the enemy, the agent disappears from the observation.
    \item When the agent wins the game by reaching the goal with the message, the goal disappears from the observation.
\end{itemize}

\paragraph{Action.} The agent can navigate the grid using five actions: \texttt{left}, \texttt{right}, \texttt{up}, \texttt{down}, and \texttt{stay}.
The agent can only interact with entities when it is in the same grid cell as the entity.
\footnote{We observed an inconsistency in the implementation with the environment description outlined in \cite{emma}.
    The agent can collide with entities even when they are not in the same grid cell.
    This is however deemed acceptable to the policy as the agent is still able to try to either go back or stay away from the other entity.
    More details can be found in this discussion: \href{https://github.com/ahjwang/messenger-emma/issues/6}{https://github.com/ahjwang/messenger-emma/issues/6}}

\subsubsection{Entities and Language Manuals}
\label{sec:entities_language_manuals}
Each game includes a language manual and an observation containing entities and a single agent.
There are twelve different entities (e.g., \texttt{airplane}, \texttt{researcher}, etc.) denoted by a fixed set of corresponding symbols that are used consistently across game instances.
For instance, symbol \includegraphics[height=1em]{figures/plane.png} for entity \texttt{plane} shown in \cref{fig:messenger_example}.
Note that the observation does not have entity names (e.g. \texttt{airplane}) and the agent must observe the entity's symbol and ground the entity's name to its corresponding symbols from the manual.

There are also three movement types for entities: \texttt{moving}, \texttt{fleeing}, and \texttt{stationary}, which describe movement trends relative to the agent’s position.
For example, a manual "heading closer and closer to where you are" describes the movement type \texttt{moving}.

For each game, the game engine assigns different roles (\texttt{enemy, goal, messenger}) and movement types (\texttt{moving, fleeing, stationary}) to a set of entities, along with the associated language manuals containing this information.
For example, "the plane fleeing from you has the classified report".
As a result, two games with the same set of entities and identical grid-world observations can have different language manuals and, consequently, different reward and transition functions.

\subsubsection{Evaluation settings}
MESSENGER provides three stages with different levels of language generalization assessment:
\paragraph{Stage 1 (S1).} This stage tests the agent's ability to ground entity names in the manual to entity symbols in the observation.
Test games offer two different levels of generalization evaluation.
First, new languages describing the same entity name using synonyms, e.g. \texttt{researcher-scholar}.
Second, new languages describing new combinations of known entities in a game, i.e. the agent has played with entities \texttt{ferry}, \texttt{plane}, \texttt{researcher} in train but not in the same game, and the agent is tasked to play with all of them in a test game.

\begin{figure}
    \centering
    \includegraphics[width=0.9\textwidth]{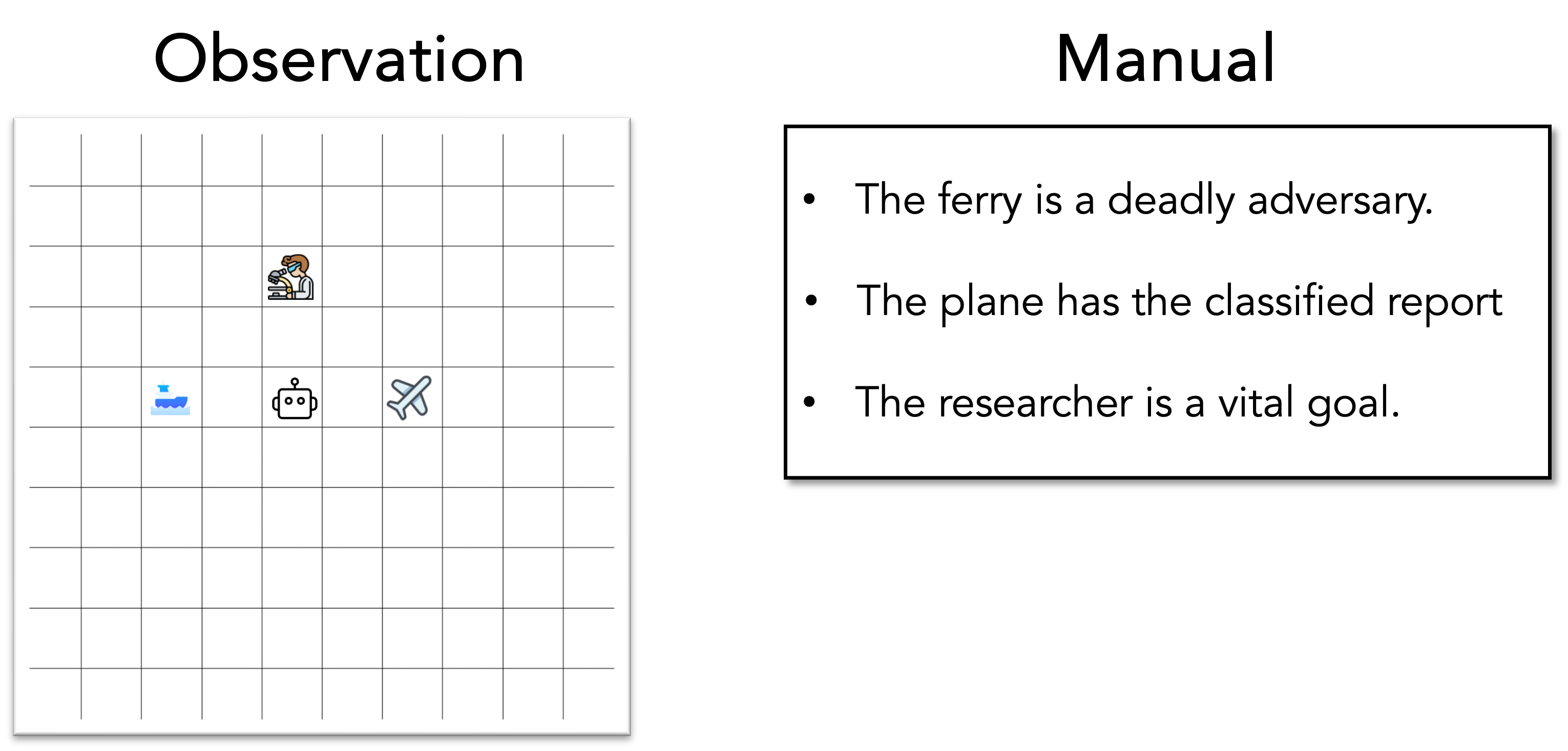}
    \caption{An example game of MESSENGER S1.
        In this game, the entity does not have message at the beginning of the game. Therefore, it goes to the messenger to retrieve the message and ends the game.
        All entities except the agent are stationary, thus the manual only describes roles associated with entity names. }
    \label{fig:s1}
\end{figure}

As shown in \cref{fig:s1}, this stage includes three entities, each with one of the three roles: \texttt{enemy}, \texttt{messenger}, and \texttt{goal}, along with their corresponding descriptions.
All entities are stationary and placed two steps away from the agent, which starts in the center of the grid.
The language descriptions only specify the entities and their roles, with no mention of movement.
The agent starts the game either with or without the \texttt{message}.

\begin{figure}
    \centering
    \includegraphics[width=\textwidth]{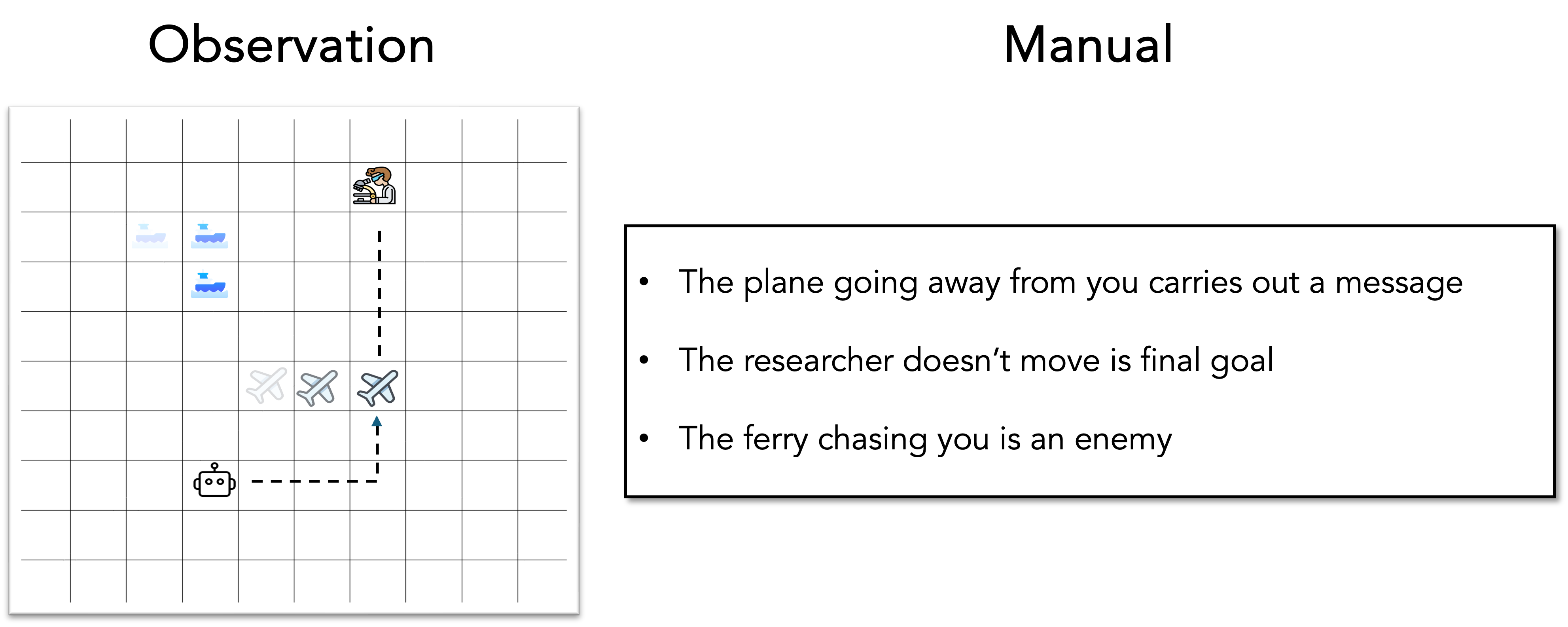}
    \caption{
        An example of a game play within a 10 × 10 grid-world from MESSENGER S2.
        The observation on the left includes three entities represented by their associated symbols: (\texttt{ferry} - \includegraphics[height=1em]{figures/ferry.png}), (\texttt{plane} - \includegraphics[height=1em]{figures/plane.png}), (\texttt{researcher} - \includegraphics[height=1em]{figures/scientist.png}) and one agent (depicted by \includegraphics[height=1em]{figures/bot.png}).
        The game involves three roles: \texttt{messenger}, \texttt{goal}, and \texttt{enemy}.
        The agent's task is to identify roles of all entities, locate the \texttt{messenger}, deliver it to the \texttt{goal}, and avoid the \texttt{enemy}.
        To achieve this objective, the agent must use the manual to infer entity roles based on their described dynamics and observed behavior.
        In the observation in the example, shaded icons indicate one possible scenario of entity locations over time.
        By observing entity movement patterns and grounding language to entities based on their according behaviors, the agent can infer the roles are assigned: \texttt{(ferry-enemy)}, \texttt{(plane-messenger)}, and \texttt{(researcher-goal)}.
        After inferring all entity roles, the agent can execute an appropriate plan to complete the task.
        The dashed line in the observation shows such a possible plan.}
    \label{fig:messenger_example_dup}
\end{figure}

\paragraph{Stage 2 (S2) and S2-dev.} As shown in \cref{fig:messenger_example_dup} \footnote{This figure is the duplication of \cref{fig:messenger_example} and is put here for the sake of reading flow.}, S2 uses the same set of entities as S1 but introduces movement dynamics: entities can now exhibit one of three movement types: \texttt{moving}, \texttt{fleeing}, or \texttt{stationary}.
The agent always starts without the \texttt{message}.
During training games, only \textit{one} movement combination is used: one \texttt{moving} (chasing), one \texttt{fleeing}, and one \texttt{stationary} entity, all of which describe how entities are moving compared to the agent.
In test games, the agent must handle scenarios where a movement type can appear multiple times, e.g. \texttt{moving-moving-fleeing}.
To examine the impact of this single-movement constraint, MESSENGER provides a different stage S2-dev, a variation of S2 that also features unseen dynamics but maintains the same movement constraint observed during training for all test games.

In addition to the capabilities demonstrated in S1, the objective of the agent in S2 is to generalize across new language featuring novel environmental dynamics.
Specifically, the agent must understand the movement descriptions to make optimal actions, but does not need to ground movement descriptions to the entities based on their observed behaviors.
This is because the agent can ground the sentences to the entities based on their names in the manual and their associated symbols in the observation.
For example, given the game in \cref{fig:messenger_example}, the agent can ground the sentence "The plane fleeing from you has the classified report" to entity symbol \includegraphics[height=1em]{figures/plane.png} based on the entity name \texttt{the plane}-to-symbol \includegraphics[height=1em]{figures/plane.png} mapping.
The agent must understand the entity's behavior to move closer to it, and it can achieve this based on the description "The plane fleeing from you", even without directly observing the behavior.

\begin{figure}
    \centering
    \includegraphics[width=\textwidth]{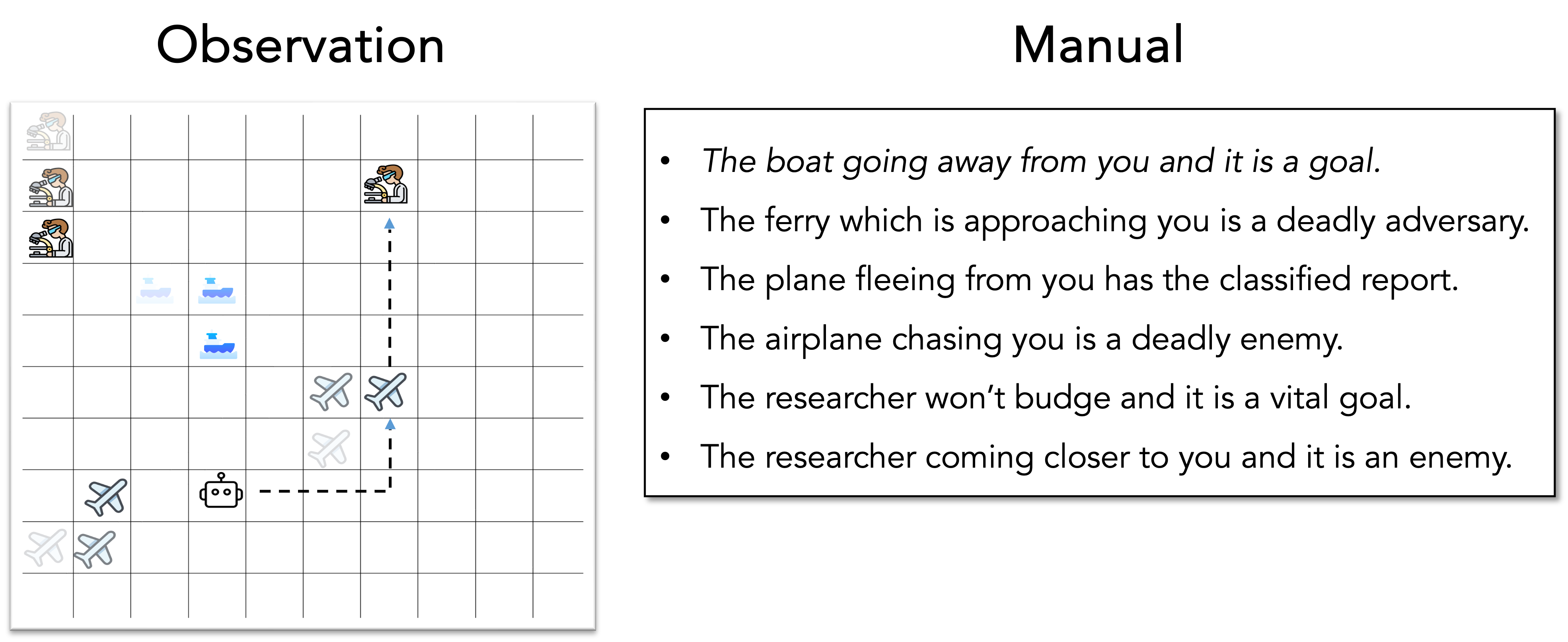}
    \caption{An example game of MESSENGER S3.
        To win the game, the agent must infer the roles of entities given the manual.
        Specifically, the same entity names (e.g. \texttt{airplane, plane} with different roles (e.g. \texttt{enemy, messenger}) must be disambiguated by their movement dynamics (e.g. \texttt{chasing, fleeing}).
        Note that we have a \textit{italicized} sentence describing an extraneous entity that is not available in the game observation.
        We also have synonyms for entity names and roles, e.g., \texttt{airplane, plane; adversary, enemy}.
        The shaded entities show possible entity locations over time and the dashed line shows a possible path for the agent to win the game. }
    \label{fig:s3}
\end{figure}

\paragraph{Stage 3 (S3).} In addition to the capabilities demonstrated in Stage 1, the objective of the agent in Stage 3 is to generalize over new language featuring new combinations of known entity movement dynamics.
Unlike in Stage 2, the agent in S3 must ground the sentences using both entity name-to-symbol mappings and observed entity behavior-to-movement description mappings.

As shown in \cref{fig:s3}, this stage includes five entities, three retain the roles of \texttt{enemy}, \texttt{messenger}, and \texttt{goal}.
The manual has six sentences featuring these five entities and one extraneous entity, which is not available in the observation.
Specifically, its referred sentence has the same name as the \texttt{enemy} entity, but is different in movement and is described as either \texttt{goal} or \texttt{messenger}.
Two additional entities are duplicates that share the same entity symbols and names of the \texttt{messenger} and \texttt{goal} accordingly, but they are assigned the role of \texttt{enemy}.
To differentiate these entities, their movement dynamics must be used.
For example, descriptions like "the fleeing enemy is the dog" versus "the chasing goal is the dog" help the agent identify the correct entities based on their behaviors.
In this case, the dog that is consistently going towards the agent can be inferred as the goal.

\subsection{Examples of different level of generalization evaluation in MESSENGER and MESSENGER-WM environment }
\label{sec:example_gen}
\begin{figure}[!htbp]

    \includegraphics[width=\textwidth]{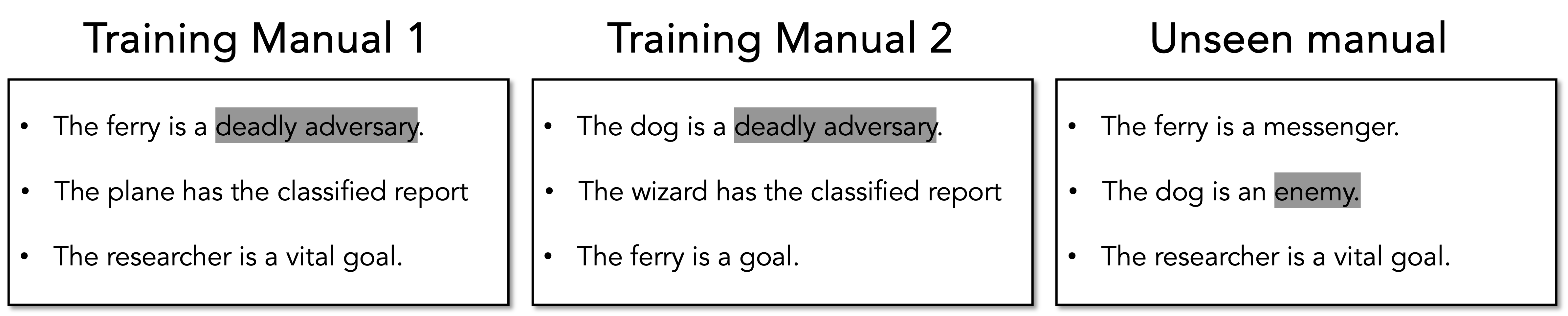}
    \caption*{(a) Novel language through synonyms and paraphrase}
    \label{fig:surface}

    \vspace{0.3cm}

    \includegraphics[width=\textwidth]{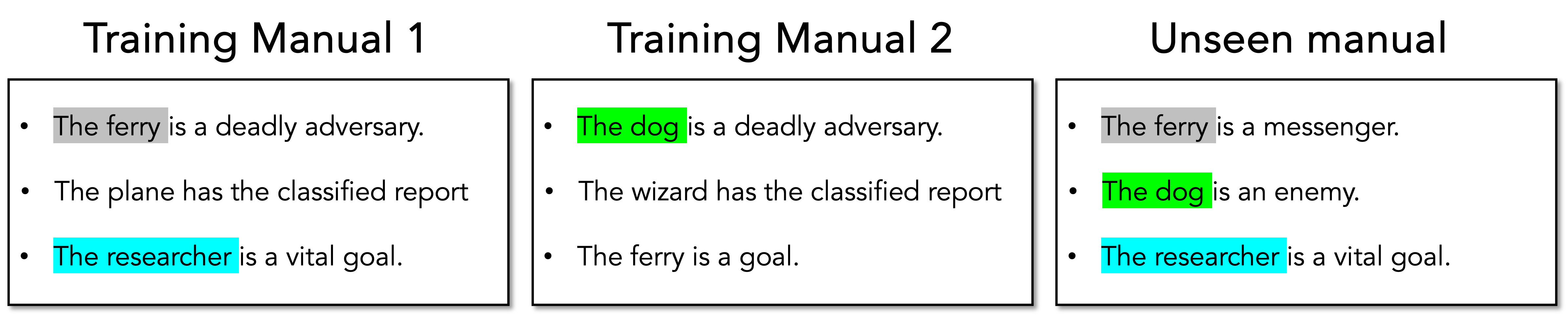}
    \caption*{(b) Novel combinations of known entities. In test games, entities never appear in the same game during training.}
    \label{fig:novelcombo}

    \vspace{0.3cm}

    \includegraphics[width=\textwidth]{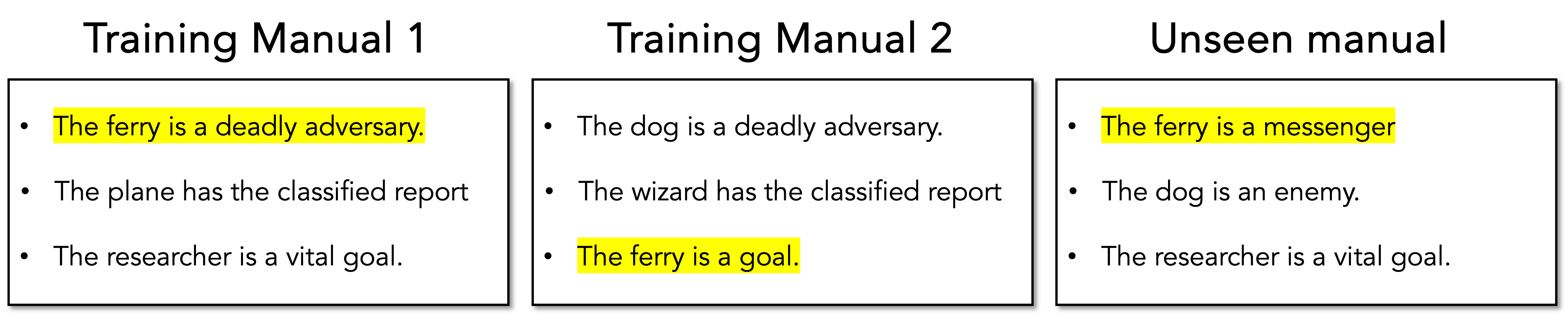}
    \caption*{(c) Novel entity-role combinations. In test games, at least one entity-role combination is unseen during training.}
    \label{fig:novel_entity_role}

    \vspace{0.3cm}

    \includegraphics[width=\textwidth]{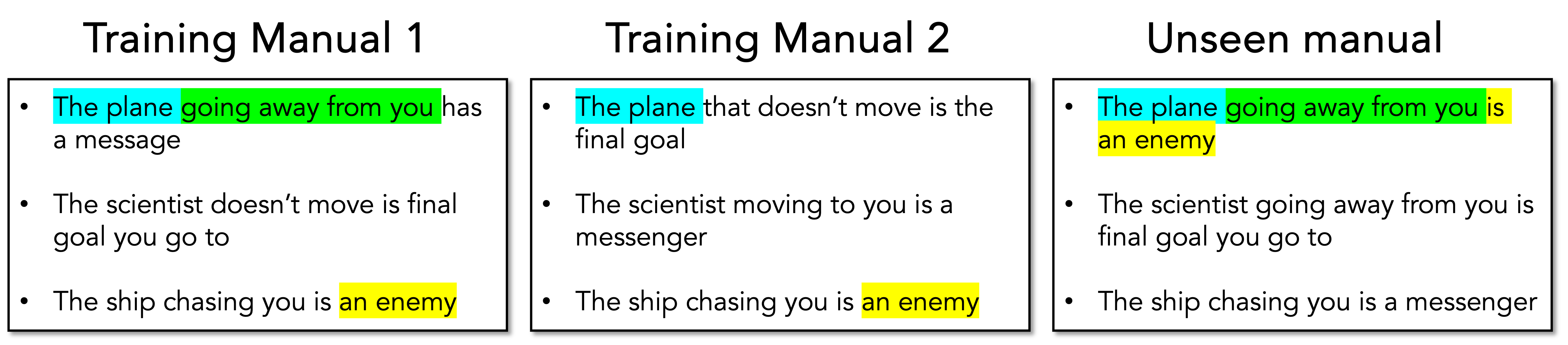}
    \caption*{(d) Novel \highlight{entityblue}{entity}-\highlight{roleyellow}{role}-\highlight{movementgreen}{movement} combinations. In test games, at least one entity-role-movement combination is unseen during training.}
    \label{fig:novel_entity_role_move}

    \vspace{0.3cm}

    \includegraphics[width=\textwidth]{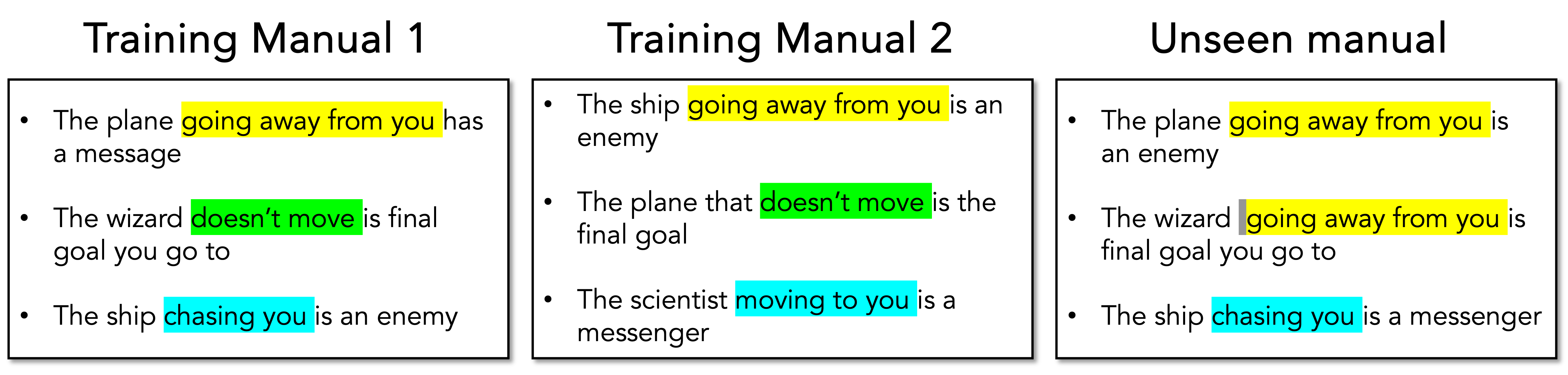}
    \caption*{(e) Novel combinations of known entity movements or novel game dynamics.
        In training games, there is only one movement combination \highlight{entityblue}{\texttt{chasing}}-\highlight{roleyellow}{\texttt{fleeing}}-\highlight{movementgreen}{\texttt{stationary}}.
        In test games, there are more than one movement combination.}
    \label{fig:novel_dynamics}

    \caption{Examples of different levels of generalization evaluation in MESSENGER and MESSENGER-WM environments.
    }
    \label{fig:generalization_examples}
\end{figure}

We illustrate different levels of generalizations, described in \cref{sec:generalization_details}, by examples in \cref{fig:generalization_examples}. For each type of generalization, for the sake of simplicity, we consider a hypothetical dataset where a training set consists of only two samples, and one test sample. The best view is in colors.

\subsection{MESSENGER-WM details}

\subsubsection{Differences between MESSENGER and MESSENGER-WM}
\label{sec:diff_wm_messenger}

Similar to MESSENGER S2, in MESSENGER-WM, there are three entities with three roles: \texttt{messenger}, \texttt{enemy}, and \texttt{goal}.
However, unlike S2 where training games only have one movement combination \texttt{chasing-fleeing-stationary}, training games in MESSENGER-WM can have more than one entity having the same movement pattern, e.g. \texttt{chasing-chasing-stationary}.
This makes generalization in S2 is more challenging than MESSENGER-WM because the agent must overcome this data bias to generalize over unseen combinations of movement patterns.

\subsubsection{Evaluation settings}
\label{sec:messenger_wm_example}

To help ground our descriptions of these settings, consider the following manual for a hypothetical test game:

\begin{quote}
    \texttt{The hound is a deadly opponent. It is towards you.} \\
    \texttt{The whale comes towards you as the secret document.} \\
    \texttt{It also has the crucial goal, the queen, and is something that cannot be moved.} \\
\end{quote}

This manual describes the entity combination: \texttt{hound, whale, queen}, with the following feature assignments: \texttt{(hound-chasing-enemy)}, \texttt{(whale-chasing-messenger)}, \texttt{(queen-stationary-goal)}.
Descriptions of each setting are as follows, based on whether the test game falls into each:

\begin{enumerate}
    \item \textbf{NewCombo:} Each game represents an unseen combination of entities.
          However, any entity-role-movement combination in this set also presents in the training games.
          In this example, the agent never sees entity combination \texttt{(hound, whale, queen)} in the same game during training, although it can see each entity individually across different games.

    \item \textbf{NewAttr:} Each game features seen combinations of entities, but at least one attribute (role, movement type, or both) for each entity is novel.
          In is example, the agent has seen entity combination  \texttt{(hound, whale, queen)} during training but each entity-role-movement assignment is new: i.e. the assignment \texttt{(hound-chasing-enemy)} is unseen but \texttt{(hound-chasing-goal)} or \texttt{(hound-fleeing-messenger)} are seen during training.

    \item \textbf{NewAll:} This setting combines the challenges of the first two.
          The combination of entities is novel, and each entity is assigned at least one new attribute.
          In this example,  entity combinations \texttt{hound, whale, queen} and all entity-role-movement (i.e. \texttt{(hound-chasing-enemy)}, \texttt{(whale-chasing-messenger)}, \texttt{(queen-stationary-goal)}) are unseen.

\end{enumerate}

\section{LED-WM details} \label{sec:led_wm_details}
Our world model LED-WM is built on the Recurrent State Space Model (RSSM) in \cite{dreamerv3}. However, we make the following modifications.
First, we find that reconstruction decoder in DreamerV3 negatively impacts policy generalization.
Therefore, we omit the decoder from DreamerV3 architecture.
Second, to improve sample efficiency, we adopt multi-step prediction for reward and continue prediction \cite{Hansen2023-cd} \cite{Peri2024-kb}.
Specifically, we rollout the latent states in the future for $H$ steps to supervise reward and continue prediction.
We replace the encoder of DreamerV3 with our LED encoder, and keep the rest of the architecture unchanged, resulting in the following components:

\begin{align}
    \raisebox{2.2em}{
        $\text{RSSM}~~\begin{cases} \hphantom{A} \\[-5pt] \hphantom{A} \\[-5pt] \hphantom{A} \end{cases}$}
    \begin{alignedat}{3}
         & \text{Sequence model:}                & \quad h_t       & = f_\phi(h_{t-1}, z_{t-1}, a_{t-1})  \\
         & \text{\textbf{LED} (\cref{sec:led}):} & \quad z_t       & \sim q_\phi(z_t \mid h_t, x_t)       \\
         & \text{Dynamics predictor:}            & \quad \hat{z}_t & \sim p_\phi(\hat{z}_t \mid h_t)      \\
         & \text{Reward predictor:}              & \quad \hat{r}_t & \sim p_\phi(\hat{r}_t \mid h_t, z_t) \\
         & \text{Continue predictor:}            & \quad \hat{c}_t & \sim p_\phi(\hat{c}_t \mid h_t, z_t) \\ \\
    \end{alignedat}
    \label{eq:led}
\end{align}

\paragraph{World model loss.} Given a sequence of observations $o_{1:T}$, actions $a_{1:T}$, rewards $r_{1:T}$ and continuation flags $c_{1:T}$ where $T$ is the horizon of a training episode, we optimize the world model parameters $\phi$ to minimize the following loss:

\begin{align}
    \mathcal{L}(\phi) \doteq \mathbb{E}_{q_\phi} \left[ \sum_{t=1}^{T} \left( \beta_{\text{pred}} \mathcal{L}_{\text{pred}}(\phi) + \beta_{\text{dyn}} \mathcal{L}_{\text{dyn}}(\phi) + \beta_{\text{rep}} \mathcal{L}_{\text{rep}}(\phi) \right) \right],
\end{align}

\noindent where $\mathcal{L}_{pred}$, $\mathcal{L}_{dyn}$, and $\mathcal{L}_{rep}$ are prediction loss, dynamics loss, and representation loss, along with their corresponding weights $\beta_{pred}$, $\beta_{dyn}$ and $\beta_{rep}$:

\begin{enumerate}
    \item Prediction loss ($\mathcal{L}_{pred}$): trains the reward predictor via symlog loss and the continue predictor via binary classification loss.
          To address the issue of slower training caused by removing the decoder and observational reconstruction loss, we adopt multi-step prediction  for reward and continue \cite{Hansen2023-cd} \cite{Peri2024-kb}, to improve sample efficiency.
          We rollout the latent states in the future for $H$ steps to supervise reward and continue prediction.
          Specifically, given a state-action trajectory over $H+1$ step $(x_t, a_t, x_{t+1}, \ldots, x_{t+H})$ associated with a sequence of rewards $r_{t:t+H}$ and continue flags $c_{t:t+H}$, we first compute $z_t$ as the posterior state from $x_t$.
          We then rollout this $z_t$ over $H$ steps to get prior states $\hat{z}_{t+1...t+H-1}$ and deterministic states $h_{t+1...t+H}$ to predict rewards $r_{t+1:t+H}$ and continue flags $c_{t:t+H}$:
          \begin{align}
              \mathcal{L}_{pred} & = \underbrace{-\ln p_\theta(r_t | z_t, h_t)}_{\text{reward loss}}
              \underbrace{-\ln p_\theta(c_t | z_t, h_t)}_{\text{continue loss}} \notag               \\
                                 & \quad \underbrace{-\sum_{k=t+1}^H
                  \lambda^{k-t-1} \left[ \ln p_\theta(r_k \mid \hat{z}_k, h_k) + \ln p_\theta(c_k \mid \hat{z}_k, h_k) \right],
              }_{\text{multi-step reward and continue loss}}                                         \\
          \end{align}

          where $\lambda=0.9$ is a discount factor when the environment is stochastic and $\lambda=1$ when the environment is deterministic.
          Recall that $\hat{z}_k$ denotes the prior stochastic state generated by the world model, without access to observation at time step $k$.

    \item Dynamics and representation loss: We adopt the dynamics and representation loss unchanged from Dreamerv3.

\end{enumerate}

\section{Training procedure}
\label{sec:training_procedure}
During world model training, we observe two challenges.
First, at the beginning of training, successful episodes in which the agent wins the game and receives positive rewards are rare.
As a result, the world model takes longer to learn from these rare instances, leading to reduced sample-efficiency in policy training.
Second, as the policy converges and produces mostly successful episodes, the replay buffer becomes dominated by these episodes.
This causes the world model to rarely encounter failed episodes where the agent loses the game and receives negative rewards, potentially harming its generalization performance.
We therefore adapt the following strategies to improve sample efficiency:

\paragraph{Prioritized Replay Buffer.}
We adopt the Prioritized Replay Buffer from \cite{Kauvar2023-pk}, where the authors propose the following prioritization strategies:

\begin{itemize}
    \item Count-based replay: Biases sampling towards recent experiences in the replay buffer.
    \item Adversarial replay: Prioritizes experiences where the world model makes incorrect predictions.
\end{itemize}

\paragraph{Balanced weights.}
We adopt a balanced weighting technique for handling class imbalance, inspired by methods used in classification tasks \cite{He2009-bj}, and apply it to world model training.
This weighting ensures that underrepresented classes contribute proportionally more to the training loss, improving sample-efficiency of the policy.

In our setting, for a given episode \( e \) with \( T \) is the episode horizon and the state-action trajectory: $\tau_e = (o_1, a_1, \ldots, o_T, a_T)$,  we define the "class" \( c_e \) as the accumulated sum of rewards for the episode:
\begin{equation}
    c_e = \sum_{t=1}^T r_t(o_t, a_t),
\end{equation}
representing different gameplay scenarios.
For example, in the MESSENGER environment, episodes fall into three classes: 1.5, -0.5, and -1.

To address the imbalance between training instances of negative classes and positive class in the replay buffer, we scale the world model loss \( \mathcal{L}_{\text{pred}} \) in each class proportional to the inverse square root of its frequency in the replay buffer.
The scaled loss is computed as:
\begin{equation}
    \mathcal{L}_{\text{pred}} = \mathcal{L}_{\text{pred}} \times \sqrt{\frac{|RB|}{\text{count}(c)}},
\end{equation}
where \( |RB| \) is the total number of episodes in the replay buffer \( RB \), and \( \text{count}(c) \) is the number of instances of class \( c \) in the replay buffer.

\paragraph{Increase throughput for replay buffer.}

In the original Dreamerv3 implementation, one trajectory with $L$ time steps of observations in the replay buffer is duplicated $L$ times, making the training data throughput inefficient.
We therefore remove this duplication to speed up the training throughput in the replay buffer, resulting in more sample-efficient.

\section{Finetune a trained policy using a trained world model}
\label{sec:finetune_algo}

\begin{algorithm}
    \DontPrintSemicolon
    \SetAlgoLined
    \SetInd{1em}{1em} %
    \SetNlSkip{1.2em}  %

    \KwIn{The trained LED-WM, the trained policy $\pi$, a test game $G$ with the first observation $obs_o$ and a language manual $L$.}
    \KwOut{A finetuned policy $\hat{\pi}$ if needed}
    \vspace{0.8em} %

    \SetKwFunction{EstimateReturn}{EstimateReturn}
    \SetKwFunction{GenerateTrajectories}{GenerateTrajectories}
    \SetKwFunction{FineTune}{FineTune}
    \SetKwProg{Fn}{Function}{:}{}

    \Fn{\EstimateReturn($\pi$, LED-WM, $obs_o$, $L$)}{
        returns = [] \;

        for \_ in range(60): \tcp*[r]{Generate synthetic test trajectories}
        \Indp
        traj = LED-WM.GenerateTrajectory($obs_o$, $L$) \;
        returns.append(sum\_rewards(traj)) \;
        \Indm

        $\hat{V}_\pi$ = mean(returns) \;
        return $\hat{V}_\pi$ \;
    }

    \vspace{0.8em} %

    \Fn{\FineTune($\pi$, LED-WM, $obs_o$, $L$)}{
    for gradient\_step in range(2000):\;
    \Indp
    trajectories = [] \;
    for \_ in range(60):\;
    \Indp
    trajectories.append(LED-WM.GenerateTrajectory($obs_o$, $L$))\;
    \Indm
    $\pi$.train(trajectories)\;

    \Indm
    return $\pi$
    }
    \vspace{0.8em} %

    \tcp{Main function: Finetune the policy $\pi$ using LED-WM}
    \Fn{PolicyFinetune(LED-WM, $\pi$, $obs_0$, L)}
    {
        $\hat{V}_\pi$ = \EstimateReturn{LED-WM, $\pi$, $obs_0$, $L$}\;
        if $\hat{V}_\pi >= thres$:\;
        \Indp
        $\hat{\pi} = \pi$\;
        \Indm
        else:\;
        \Indp
        $\hat{\pi}$ = \FineTune{$\pi$, LED-WM, $G$, $obs_0$, $L$}\;
        \Indm
        return $\hat{\pi}$\;
    }
    \caption{Policy Finetune with LED-WM}
    \label{algo:finetune}
\end{algorithm}

\end{appendices}

\newpage
\section*{NeurIPS Paper Checklist}

\begin{enumerate}

    \item {\bf Claims}
    \item[] Question: Do the main claims made in the abstract and introduction accurately reflect the paper's contributions and scope?
    \item[] Answer: \answerYes{} %
    \item[] Justification: Our claim about policy generalization and world model generalization in \cref{sec:introduction} are reflected in our experiment results in \cref{sec:experiment}.

    \item {\bf Limitations}
    \item[] Question: Does the paper discuss the limitations of the work performed by the authors?
    \item[] Answer: \answerYes{} %
    \item[] Justification: We discuss our limitations in policy generalization where CRL \cite{crl} outperforms LED-WM in S2 in \cref{sec:experiment}.

    \item {\bf Theory assumptions and proofs}
    \item[] Question: For each theoretical result, does the paper provide the full set of assumptions and a complete (and correct) proof?
    \item[] Answer: \answerYes{} %
    \item[] Justification: In \cref{sec:introduction}, we stated that our assumption is that our observation is symbolic and confied to a discrete grid-world.

    \item {\bf Experimental result reproducibility}
    \item[] Question: Does the paper fully disclose all the information needed to reproduce the main experimental results of the paper to the extent that it affects the main claims and/or conclusions of the paper (regardless of whether the code and data are provided or not)?
    \item[] Answer: \answerYes{} %
    \item[] Justification: The paper provides the code and hyperparameters for training the world model and the policy in \cref{sec:experiment} and \cref{sec:training_details}.

    \item {\bf Open access to data and code}
    \item[] Question: Does the paper provide open access to the data and code, with sufficient instructions to faithfully reproduce the main experimental results, as described in supplemental material?
    \item[] Answer: \answerYes{} %
    \item[] Justification: We provide the code and hyperparameters for training the world model and the policy in \cref{sec:experiment} and \cref{sec:training_details}.

    \item {\bf Experimental setting/details}
    \item[] Question: Does the paper specify all the training and test details (e.g., data splits, hyperparameters, how they were chosen, type of optimizer, etc.) necessary to understand the results?
    \item[] Answer: \answerYes{} %
    \item[] Justification: We adopt MESSENGER and MESSENGER-WM environments which follow their standard train/dev/test split. We provide our code and necessary hyperparameters in \cref{sec:training_details}.

    \item {\bf Experiment statistical significance}
    \item[] Question: Does the paper report error bars suitably and correctly defined or other appropriate information about the statistical significance of the experiments?
    \item[] Answer: \answerYes{} %
    \item[] Justification:
          We use Wilcoxon signed-rank \cite{Wilcoxon1945-sa}, bootstrap sampling \cite{Efron1979-wz}, and hierarchical bootstrap sampling \cite{Davison2013-hr} to do statistical tests for policy finetune results in \cref{tab:finetune_messenger} in S2-dev.

    \item {\bf Experiments compute resources}
    \item[] Question: For each experiment, does the paper provide sufficient information on the computer resources (type of compute workers, memory, time of execution) needed to reproduce the experiments?
    \item[] Answer: \answerYes{} %
    \item[] Justification: We provide training time and our used GPU information in \cref{tab:env_hparams}.

    \item {\bf Code of ethics}
    \item[] Question: Does the research conducted in the paper conform, in every respect, with the NeurIPS Code of Ethics \url{https://neurips.cc/public/EthicsGuidelines}?
    \item[] Answer: \answerYes{} %
    \item[] Justification: The authors have reviewed the NeurIPS Code of Ethics and the paper conforms to the code of ethics.

    \item {\bf Broader impacts}
    \item[] Question: Does the paper discuss both potential positive societal impacts and negative societal impacts of the work performed?
    \item[] Answer: \answerNA{} %
    \item[] Justification: We do not have any potential positive or negative societal impacts.

    \item {\bf Safeguards}
    \item[] Question: Does the paper describe safeguards that have been put in place for responsible release of data or models that have a high risk for misuse (e.g., pretrained language models, image generators, or scraped datasets)?
    \item[] Answer: \answerNA{} %
    \item[] Justification: We do not have any data or models that have a high risk for misuse.

    \item {\bf Licenses for existing assets}
    \item[] Question: Are the creators or original owners of assets (e.g., code, data, models), used in the paper, properly credited and are the license and terms of use explicitly mentioned and properly respected?
    \item[] Answer: \answerYes{} %
    \item[] Justification: We cited the original papers for the environments and other baselines.

    \item {\bf New assets}
    \item[] Question: Are new assets introduced in the paper well documented and is the documentation provided alongside the assets?
    \item[] Answer: \answerYes{} %
    \item[] Justification: We will release the code and instructionsfor our experiments.

    \item {\bf Crowdsourcing and research with human subjects}
    \item[] Question: For crowdsourcing experiments and research with human subjects, does the paper include the full text of instructions given to participants and screenshots, if applicable, as well as details about compensation (if any)?
    \item[] Answer: \answerNA{} %
    \item[] Justification: We do not involve crowdsourcing nor research with human subjects.

    \item {\bf Institutional review board (IRB) approvals or equivalent for research with human subjects}
    \item[] Question: Does the paper describe potential risks incurred by study participants, whether such risks were disclosed to the subjects, and whether Institutional Review Board (IRB) approvals (or an equivalent approval/review based on the requirements of your country or institution) were obtained?
    \item[] Answer: \answerNA{} %
    \item[] Justification: We do not involve research with human subjects.

    \item {\bf Declaration of LLM usage}
    \item[] Question: Does the paper describe the usage of LLMs if it is an important, original, or non-standard component of the core methods in this research? Note that if the LLM is used only for writing, editing, or formatting purposes and does not impact the core methodology, scientific rigorousness, or originality of the research, declaration is not required.
    \item[] Answer: \answerNA{} %
    \item[] Justification: we only use LLM for editing purposes.

\end{enumerate}

\end{document}